# Vibration suppression of a state-of-the-art wafer gripper

M. El Ajjaj[1], M.B. Kaczmarek[1], M.A.C.C. van den Hurk[2], S.H. HosseinNia[1]

[1]Delft University of Technology (Department: Precision and Microsystems), The Netherlands
[2]VDL Enabling Technologies Technology and Development, The Netherlands

M.B.Kaczmarek@tudelft.nl

**Abstract**

The demand for thin lightweight structures has been ever increasing, which leads to new challenges on vibration control technology. This especially holds for the semiconductor industry where high standards are set when it comes to speed, precision and alignment accuracy. Flexible robot end-effectors, such as wafer grippers, tend to be prone to vibrations due to their lightweight design and low thickness. The added challenge is that the contact between the wafer and the gripper tends to be friction-based. When vibrations are not attenuated for these thin end-effectors, the wafer can slip, break or even get lost in the lithography machines. Current solutions have been to increase the thickness of the grippers as much as possible within budget or to add viscoelastic material to the gripper. However, these solutions only work up until a certain point and tend to provide limited damping. Smart materials, which are characterized by their coupling between different physical domains, are currently being researched to provide new methods to contribute to vibration attenuation [1]. The most common smart materials for the implementation in thin structures are: piezoelectric materials[2]–[7], shape memory alloys[8]–[11], thermo-elastomers[2], [12], and electro-active polymers[13], [14]. These smart materials can be integrated in a so called 'smart structure', that can reduce structural vibrations through the high degree of integration of sensors, actuators and an appropriate control system [15]. Piezoelectric materials tend to be the most commonly and extensively used smart material[16]. Piezoelectric materials are materials that produce an electrical charge when a mechanical stress is applied [17]. The inverse holds as well, where when a voltage is applied to the transducer.

In this paper the implementation of piezoelectrics to a state-of-the-art wafer gripper is investigated. The objective is to propose and validate a solution method, which includes a mechanical design and control system, to achieve at least 5% damping for two eigenmodes of a wafer gripper. This objective serves as a 'proof of concept' to show the possibilities of implementing a state-of-the-art damping method to an industrial application, which in turn can be used to dampen different thin structures. The coupling relation between the piezoelectrics and their host structure were used to design the placement of the piezoelectric patches, together with modal analysis data of the a state-of-the-art wafer gripper. This data had been measured through an experimental setup. Active damping has been succesfully implemented onto the wafer gripper where positive position feedback (PPF) is used as a control algorithm to dampen two eigenmodes.

**Electromechanical coupling factor**

The electromechanical coupling factor serves as a measure to show the efficiency for which the mechanical energy is converted to electrical energy, and vice versa. This measure can be used to determine how and where to implement piezoelectrics onto a host structure. The effective coupling between a piezoelectric transducer and its host structure, in the case a wafer gripper, can be determined analytically through the admittance of a piezoelectric transducer that is 'perfectly' attached to its host structure. This means that the influence of the bonding layer between the structure and the piezoelectric transducer is left out of scope. It is assumed that admittance here is defined as the charge over voltage and indicates the amount of current that can flow through the system [18]. Another assumption is that the host structure behaves like a beam and that the piezoelectric transducer is much thinner than the thickness of the host structure.

The fromula for the coupling factor [16]

$$K_i^2 = \frac{\kappa_{31}^2}{(1-\kappa_{31}^2)} \frac{\Delta \theta_i^2}{\mu_i \omega_i^2} \frac{E_p b_p t_p z_m^2}{l} \cong \frac{\Omega_i^2 - \omega_i^2}{\omega_i^2} \tag{1}$$

can be divided into three parts. The first fraction is composed solely of material parameters of the piezoelectric transducer, where $k_{31}^2 = \frac{d_{31}^2}{s^E \epsilon^T}$ is the electromechanical coupling factor of the piezoelectric material. The second fraction depends on the modal behavior of the host structure, where $\Delta \theta_i^2$ is the difference of the slope of the mode shapes of the host structure at the beginning and end of the piezoelectric patch, $\mu_i$ is the modal mass of the host structure, and $\omega_i$ $(rad/s)$ is its eigenfrequency. The third fraction contains design parameters for the selected piezoelectric transducer, where $E_p$ $(Pa)$ is the Young's Modulus of the material, $b_p$ $(m)$ is the width of the piezoelectric patch, $t_p$ $(m)$ is the thickness, $z_m$ $(m)$ is the distance from the piezoelectric patch to the mid-axis of the host structure and $l(m)$ is the length of the piezoelectric transducer [16]. The value for the coupling factor can also be found experimentally by measuring the natural frequency of the structure for open-circuited electrodes $\Omega_i$ $(rad/s)$ and short-circuited electrodes $\omega_i$ $(rad/s)$.

Equation 1 can serve as a back of the envelope method for determining an optimal design for piezoelectric bending patches on a host structure. Moreover, this coupling factor relates to the maximum achievable modal damping, since it shows the measure of how much mechanical vibration energy can be converted into electrical energy, and vice versa. The higher the coupling factor, the more damping can be achieved.

**Experimental setup**

The experimental setup is presented in Figure 1. The wafer gripper attached to an aluminum clamp is suspended by an elastic cord from a steel frame. The gripper is connected via a thin strut to a shaker (Brüel & Kjaer Vibration Exciter type 4809). This shaker excites the clamp of the gripper, and since the clamp has a considerable larger mass than the thin wafer gripper, the wafer gripper can be considered to be clamped in this configuration. The dynamics of the gripper are of interest and therefore the chosen orientation of the gripper does not matter as gravity should not influence the results. A chirp signal is sent to the shaker, after which the displacement is measured at a pre-specified location on the wafer gripper.

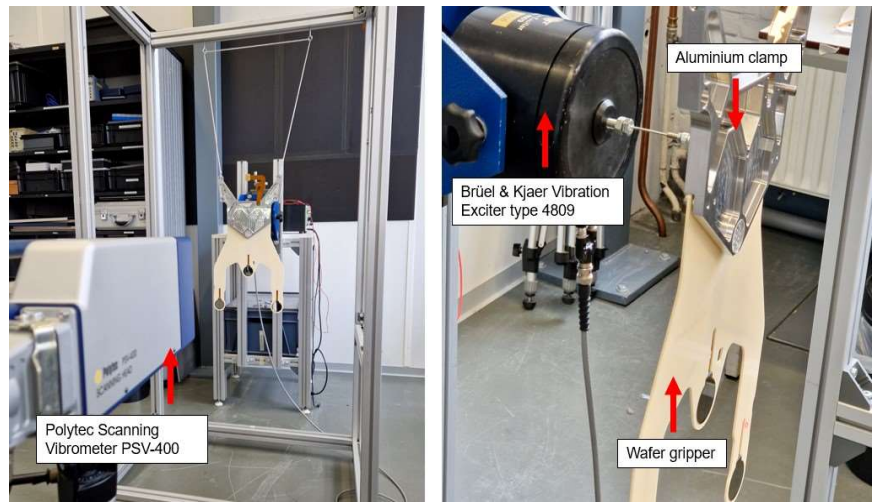

Figure 1: Experimental setup

This measurement is done through the use of a laser doppler scanning vibrometer (Polytec Scanning Vibrometer PSV-400). The vibrometer measures a high frequency signal which can be chosen to be decoded by different velocity or displacement decoders. The Polytec OFV-5000 Controller that comes with its own commercial software package, PSV-E-401 Junction Box, PSV-I-400 Scanning Head and PSV-W-401 PC are used to complete the scanning vibrometry setup.

**Modal Analysis**

Before implementing piezoelectric patches onto the gripper, the modal behaviour of the wafer gripper is analysed. Firstly, a single point measurement is done using the laser Doppler vibrometer, where the tip velocity is measured and the wafer gripper is excited by the shaker. To improve the quality of the measurement a small piece of reflective tape (ifm electronic E21015) is used to increase the reflectivity of the surface.

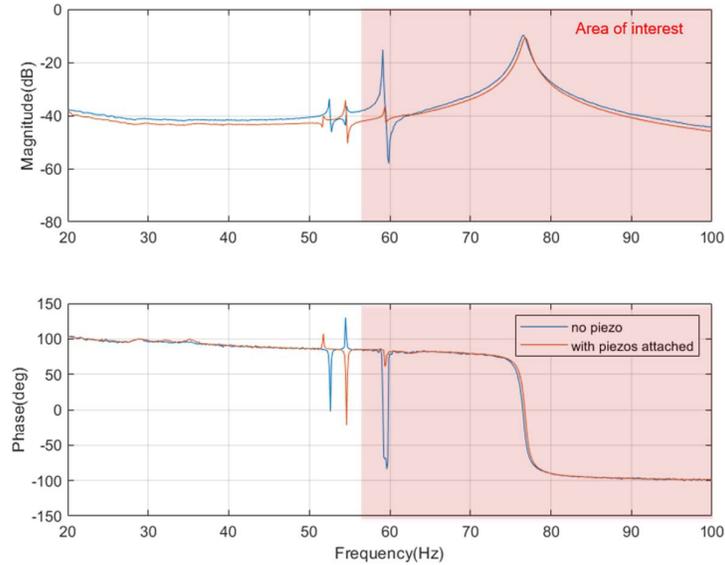

*Figure 2: Frequency response of the wafer gripper with shaker excitation as an input and tip displacement as an output with and without piezoelectric patches attached.*

The results of single-point identification are presented with the blue line in Figure 2. It can be seen that the two modes with the largest magnitude occur at around 58Hz and 76Hz. Therefore, the decision was made to target these modes for dampening the system. The displacement data for the mode shapes is presented in Figure 3. The measurements were done along a line, as presented in Figure 4.

**Design of a damped flexible end-effector**

In order to allow for a proper implementation of the piezoelectric transducer onto the wafer gripper, the type of actuator needs to be selected. Assuming a 'perfect' bonding layer, Equation is used to choose the piezoelectric bending actuator such that the coupling between the transducer and host structure is optimal. Looking at commercially available piezoelectric patches and their dimensions, the patches that are best suited for actuation are the P-876.A15 DuraAct Patch Transducers from Physik Instrumente (PI). Two patches will be used for each of the fingers of the wafer gripper. For the sensing patches the P-876.SP1 DuraAct Patch Transducers from Physik Instrumente (PI) are used.

Since the dimensions of the piezoelectric patches, and the modal behavior of the gripper is now known, the placement of the transducers can be determined. Looking at Equation 1, the only variable that is related to the placement of the piezoelectrics that remains unknown is $\Delta\theta_i^2$. This value indicates the difference in the slope of the mode shape at the beginning of the piezoelectric patch ($\theta_1$) and the end of the patch ($\theta_2$). Since the other parameters are constants, the coupling factor can be set proportional to the difference in the slopes of the mode shape $K_i \propto (\theta_1 - \theta_2)$. The optimal placement along the measurement lines is presented in Figure 4.

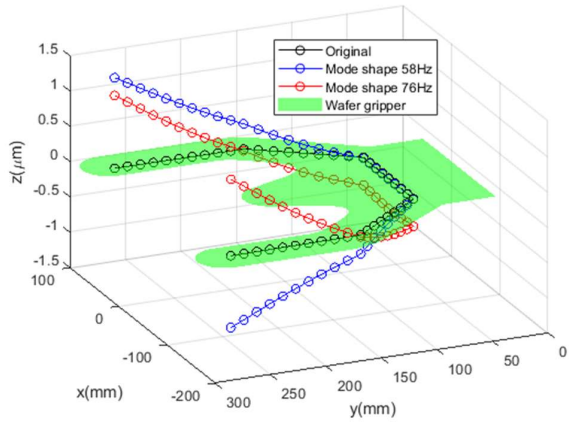
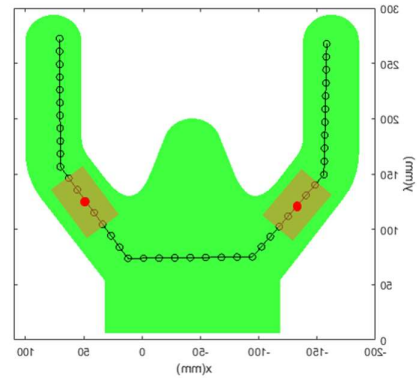

Figure 3: Mode shape displacement plot of wafer gripper

Figure 4: Geometry of wafer gripper with optimal piezopatch location

**Active setup**

The complete active dampig setup is presented in Figure 5. The piezoelectric transducers are placed at the determined locations. The gripper is covered in reflective tape, making sure to leave enough space for the piezoelectric patches and the wiring. The influence of the piezos and the reflective have been checked. Frequency responses of the gripper with and without piezoelctric patches attached are compared in Figure 2. The peak at 58Hz has been significantly reduced and does not have to be attenuated actively. The peak at around 76Hz has slightly shifted, but remains of roughly the same magnitude. Therefore, the focus will be to actively dampen the peak at around 76Hz with the use of a PPF controller.

The piezoelectric actuators are each connected to a voltage amplifier that in turn are connected to a voltage source and a micro controller with a Simulink interface. On the backside of the gripper the piezoelectric sensors (P-876.SP1 DuraAct Patch Transducers) are attached at the same location as the actuators to ensure collocation.

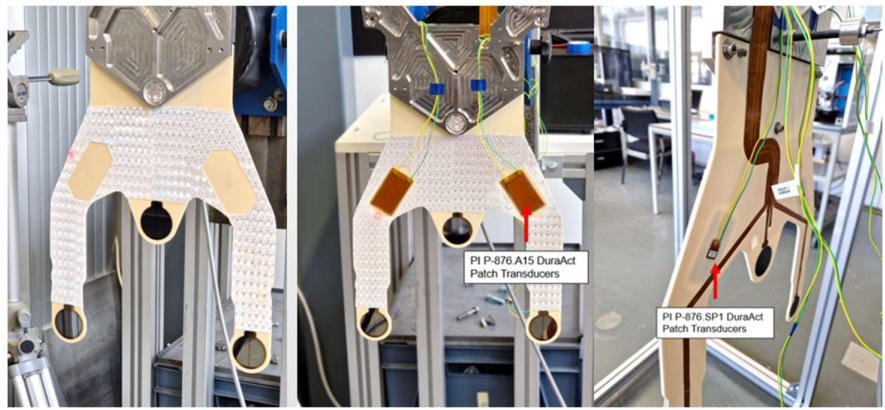

Figure 3:The wafer gripper with and without piezoelectric patches

**Results**

The PPF controller was implemented by setting the eigenfrequency at 76.7Hz. The damping ratio was set at 0.3, which is a typical value used for PPF controllers [16]. The gain was slowly increased to see the effect of the PPF controller on the damping performance of the system. In Table 1 the damping performance for different gains is presented.

*Table 1: Damping values for different PPF gains*

| Gain | Q-factor | Damping (%) |
|---|---|---|
| 0.1 | 22.9 | 2.2 |
| 0.2 | 17.6 | 2.8 |
| 0.3 | 15.7 | 3.2 |
| 0.4 | 15.2 | 3.3 |
| 0.5 | 15.2 | 3.3 |

Figure 6 shows the measured transfer functions of the wafer gripper. The magnitude of the peak decreases with an increased gain. The gain could not be further increased past 0.5, since for higher values the system became unstable.

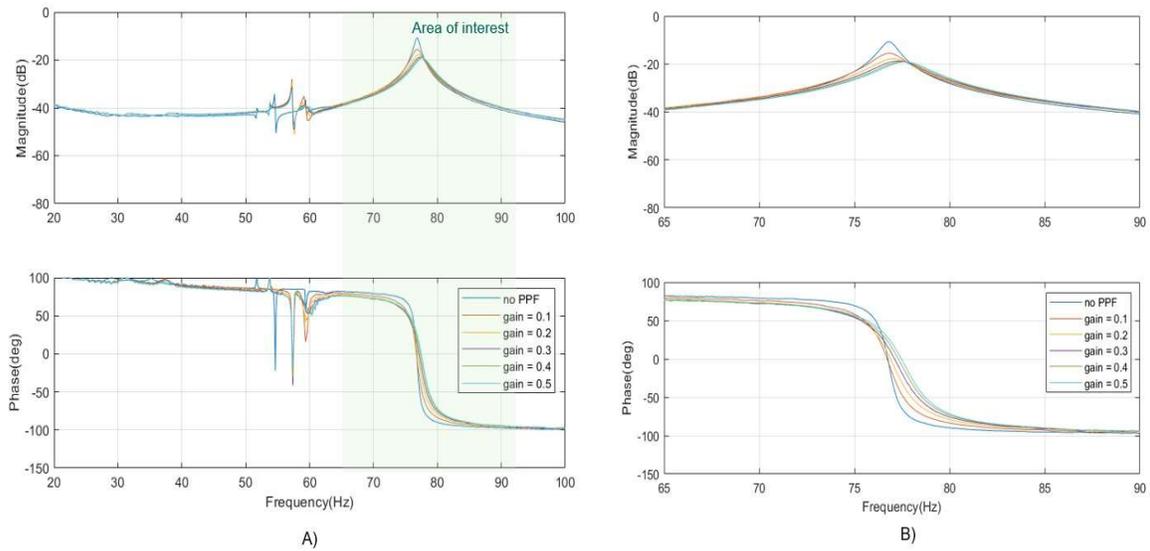

*Figure 6: Bode plots for a system with a implemented PPF controller: A) Entire plot B) Zoomed-in plot showing the eigenmode*

**Discussion and conclusion**

A design has been proposed to effectively dampen two eigenmodes of a state-of-the-art wafer gripper end-effector. Several experiments have been performed to check the modal behavior of the plant. Using this data, a design has been proposed to perform active vibration control. The eigenmode at 76.7Hz is being dampened through the use of active control and a clear reduction of the resonance peak is obtained. There is a slight shift of the resonance peak to the lower frequencies. The gain has not been further increased beyond a gain of 0.5, because the system became unstable when doing so, which prevented the measurement from being performed. This could have several possible reasons. It could be that something in the experimental setup causes this behavior, for example noise in the voltage amplifiers that cause instability. However, it could also be that there are some unknown dynamics in the system.

The goal to design a system to achieve at least 5% of modal damping for two eigenmodes has not been satisfied. However, for the second mode of interest at 76Hz, the maximum modal damping of 3.3% was achieved with a simple PPF controller with a limited gain for which the piezoelectric transducers where placed according to a simplified line optimization. This shows that the concept of dampening very stiff and thin end-effectors, such as the wafer gripper, is promising. Additionally, the possibility exists that if we optimize the placement along the surface of the gripper, there might be another better solution to the optimal placement of the piezoelectric patches. Application of the piezoelectric patches dampened the first mode of interest at 58Hz, solely because of their placement. The reason why this happens and how the placement is related to this added damping should be further investigated.

Lastly, it can be noted that the scope of the research is set at the modes of interest at 58Hz and 76Hz, but there is some change in behavior happening for the lower frequency modes (modes below 58Hz). These modes were left out of scope as the focus was set on the two largest modes. For future work the behavior that is seen for the lower frequency modes could be further investigated

# References

[1] J. A. B. Gripp and D. A. Rade, "Vibration and noise control using shunted piezoelectric transducers: A review," *Mechanical Systems and Signal Processing*, vol. 112. Academic Press, pp. 359–383, Nov. 01, 2018. doi: 10.1016/j.ymssp.2018.04.041.
[2] R. M. Syriac, A. B. Bhasi, and Y. V. K. S. Rao, "A review on characteristics and recent advances in piezoelectric thermoset composites," *AIMS Materials Science*, vol. 7, no. 6. AIMS Press, pp. 772–787, 2020. doi: 10.3934/MATERSCI.2020.6.772.
[3] Z. cheng Qiu, J. da Han, X. min Zhang, Y. chao Wang, and Z. wei Wu, "Active vibration control of a flexible beam using a non-collocated acceleration sensor and piezoelectric patch actuator," *J Sound Vib*, vol. 326, no. 3–5, pp. 438–455, Oct. 2009, doi: 10.1016/j.jsv.2009.05.034.
[4] S. S. Heganna and J. J. Joglekar, "Active Vibration Control of Smart Structure Using PZT Patches," in *Procedia Computer Science*, 2016, vol. 89, pp. 710–715. doi: 10.1016/j.procs.2016.06.040.
[5] J. Tuma, R. Strambersky, and V. Pavelka, "Modeling the Use of the Patch Piezo-Actuators for Active Vibration Control," Jun. 2021, pp. 1–4. doi: 10.1109/iccc51557.2021.9454636.
[6] P. Shivashankar and S. Gopalakrishnan, "Review on the use of piezoelectric materials for active vibration, noise, and flow control," *Smart Materials and Structures*, vol. 29, no. 5. Institute of Physics Publishing, May 01, 2020. doi: 10.1088/1361-665X/ab7541.
[7] Q. Zheng, H. Richter, and Z. Gao, "Active disturbance rejection control for piezoelectric BEAM," *Asian J Control*, vol. 16, no. 6, pp. 1612–1622, Nov. 2014, doi: 10.1002/asjc.854.
[8] C. Cho and Q. Pan, "Damping Property of Shape Memory Alloys Mechanical properties of CFRP View project High performance automotive brake disk development View project DAMPING PROPERTY OF SHAPE MEMORY ALLOYS," 2008. [Online]. Available: https://www.researchgate.net/publication/281792397
[9] H. Janocha, "Actuators in Adaptronics," in *Adaptronics and Smart structures*, Springer, 2007, pp. 95–300. doi: 10.1007/978-3-540-71967-0_6.

[10] D. C. Lagoudas, *Shape Memory Alloys*, vol. 1. Boston, MA: Springer US, 2008. doi: 10.1007/978-0-387-47685-8.
[11] A. R. Damanpack, M. Bodaghi, M. M. Aghdam, and M. Shakeri, "On the vibration control capability of shape memory alloy composite beams," *Compos Struct*, vol. 110, no. 1, pp. 325–334, Apr. 2014, doi: 10.1016/j.compstruct.2013.12.002.
[12] Y. Sun, D. Fang, and A. K. Soh, "Thermoelastic damping in micro-beam resonators," *Int J Solids Struct*, vol. 43, no. 10, pp. 3213–3229, May 2006, doi: 10.1016/j.ijsolstr.2005.08.011.
[13] C. Graf and J. Maas, "Electroactive polymer devices for active vibration damping," in *Electroactive Polymer Actuators and Devices (EAPAD) 2011*, Mar. 2011, vol. 7976, p. 79762I. doi: 10.1117/12.879934.
[14] K. Wolf, T. Röglin, F. Haase, T. Finnberg, and B. Steinhoff, "An electroactive polymer based concept for vibration reduction via adaptive supports," in *Electroactive Polymer Actuators and Devices (EAPAD) 2008*, Mar. 2008, vol. 6927, p. 69271F. doi: 10.1117/12.776294.
[15] S. Li, S. Ochs, E. Slomski, and T. Melz, "Design of control concepts for a smart beam structure with sensitivity analysis of the system," in *Computational Methods in Applied Sciences*, 2017, vol. 43, pp. 115–132. doi: 10.1007/978-3-319-44507-6_6.
[16] A. Preumont, *Vibration Control of Active Structures: An Introduction*, 4th ed. Springer, 2018. doi: 10.1007/978-3-319-72296-2.
[17] R. S. O. Moheimani and A. J. Fleming, *Piezoelectric Transducers for Vibration Control and Damping*. Springer, 2006.
[18] M. Abdus, S. Quazi, and M. Rahman, *Fundamentals of Electrical Circuit Analysis*. Springer, 2018.